\documentclass[11pt]{article}

\usepackage[]{ACL2023}

\usepackage{times}
\usepackage{latexsym}

\usepackage[T1]{fontenc}

\usepackage[utf8]{inputenc}

\usepackage{microtype}

\usepackage{inconsolata}

\usepackage{todonotes}
\usepackage{amsmath}
\usepackage{booktabs}
\usepackage{CJKutf8}

\usepackage[ruled,vlined,linesnumbered,noresetcount]{algorithm2e}
\usepackage{xurl}

\title{Recursive numeral systems are highly regular and easy to process}

\author{
  Ponrawee Prasertsom\textsuperscript{1} \quad 
  Andrea Silvi\textsuperscript{2}
  \quad 
  Jennifer Culbertson\textsuperscript{1} \\
  \textbf{Devdatt Dubhashi\textsuperscript{2}} \quad 
  \textbf{Moa Johansson\textsuperscript{2}} \quad 
 \textbf{Kenny Smith\textsuperscript{1}}\vspace{0.2cm}
  \\ 
  \textsuperscript{1}University of Edinburgh \quad
  \textsuperscript{2}Chalmers University of Technology and Gothenburg University 
  \\
   \texttt{\{ponrawee.prasertsom, jennifer.culbertson, kenny.smith\}@ed.ac.uk}\\
   \texttt{\{silvi, moa.johansson, dubhashi\}@chalmers.se}
}

\begin{document}
\maketitle

\begin{abstract}
 Much recent work has shown how cross-linguistic variation is constrained by competing pressures from efficient communication.
 However, little attention has been paid to the role of the systematicity of forms (\textit{regularity}), a key property of natural language.
 Here, we demonstrate the importance of regularity in explaining the shape of linguistic systems by looking at recursive numeral systems.
 Previous work has argued that these systems optimise the trade-off between lexicon size and average morphosyntatic complexity \citep{denic2024recursive}. 
 However, showing that \textit{only} natural-language-like systems optimise this trade-off has proven elusive, and existing solutions rely on ad-hoc constraints to rule out unnatural systems \citep{yang2025re}. 
 Drawing on the Minimum Description Length (MDL) approach, we argue that recursive numeral systems are better viewed as efficient with regard to their regularity and processing complexity. 
We show that our MDL-based measures of regularity and processing complexity better capture the key differences between attested, natural systems and theoretically possible ones, including ``optimal'' recursive numeral systems from previous work, and that the ad-hoc constraints naturally follow from regularity.
Our approach highlights the need to incorporate regularity across sets of forms in studies attempting to measure efficiency in language. 

\end{abstract}

\section{Introduction} 
One key question in the study of language concerns \textit{constrained variation}: why languages are the way they are; what cognitive and communicative constraints shape linguistic systems? Recent research suggests that linguistic systems are shaped by efficient communication, such that natural systems optimise the trade-off between simplicity and informativeness in various domains \citep{gibson2019efficiency}, from kinship terms \citep{kemp2012kinship}, colour terms \citep{regier2015word,zaslavsky2019col}, grammatical categories \citep{mollica2021forms},  indefinite pronouns \citep{denicIndefinitePronounsOptimize2022a}, boolean connectives \citep{uegaki_informativenesscomplexity_2022} to spatial demonstratives  \citep{chen2023information}.
For example, colour-naming systems tend to either partition the space of colour referents into few large, connected categories (e.g., \textit{dark} vs. \textit{light} colours), or into many small specific categories (e.g., \textit{red, green, blue, yellow, brown}, ...). The former systems are simple (i.e., they have very few categories) but they are uninformative (i.e., they do not allow precise communication about colours), whereas the terms in the latter systems are highly informative but the systems are complex \citep{regier2015word}.

While the simplicity-informativeness trade-off has been successful as an explanation for the way languages are, most work in this area does not consider the systematicity of the forms, or form-meaning mappings, in their efficiency analyses.
In other words, they do not consider that natural languages are largely \textit{regular}; they conform to rule-like patterns expressible by formal grammars 
\citep{chomsky2002syntactic,jackendoff2002foundations}.
This does not present a serious issue in most previous work, as the category labels in the domains considered, such as basic colour terms, often do not overlap in form. 
However, a more complete theory of constrained variation would necessarily have to consider the regularity that is prevalent in language.

Here, we present a first step towards incorporating regularity in a study of efficiency in language. 
We look at \textit{recursive numeral systems} (i.e., counting systems) as a case study.
Previous work on the constrained variation of recursive numeral systems considered trade-offs that are explanatorily insufficient, and/or failed to consider the systematicity of the forms, treating them as unanalysable.
\citet{xu2020numeral} initially investigated the simplicity-informativeness trade-off in the domain of numeral systems. 
They found that while approximate numeral systems, which have limited precision (e.g., \textit{a few} vs. \textit{many}), followed the trade-off, recursive numeral systems like the English decimal system (\textit{one}, \textit{two}, \ldots{} \textit{ninety}, \textit{ninety-one}, \ldots{}) are maximally informative in that any integer can be precisely expressed.
Since informativeness is fixed, \citet{denic2024recursive} (henceforth, D\&S) argued that the relevant trade-off shaping natural recursive numeral systems is instead between lexicon size
and the average length of the numerals.
However, as reviewed in Section 2, the optimal numeral systems under this trade-off included not only human-like systems, but also unnatural, highly irregular ones.
\citet{yang2025re} (Y\&R) attempted to resolve this with ad-hoc constraints that artificially restrict the space of possible systems to be human-like.

In this paper, we argue instead that the key characteristic of human-like recursive numeral systems is how regular they are (\textit{regularity}), and how easy it is to process them (\textit{processing complexity}).
We leverage a precise formal definition of regularity from formal language and complexity theory to demarcate natural language systems from other theoretically possible ones. 
Our empirical contributions are:
\begin{itemize}
    \item We introduce regularity and processing complexity as the key characteristics of recursive numeral systems (Section \ref{sec:irregularityandproccomp}),
    \item We demonstrate that human-like systems are significantly more regular and easier to process than other theoretically possible systems (Section \ref{sec:applying}), 
    \item We show that our measures capture the difference between natural systems and unattested ones regardless of assumptions about the prior over numbers (Section \ref{sec:prior}).
\end{itemize}

The code and data for our analyses are available at \url{https://github.com/ponrawee/regularity-processing-complexity-recursive-numerals}.

\section{Prior work on efficiency of recursive numeral systems}
\label{sec:literature}
D\&S proposed that recursive numeral systems optimise the trade-off between \textit{lexicon size} and \textit{average morphosyntactic complexity}. 
Lexicon size is the number of meanings encoded by single morphemes (e.g. in Mandarin this includes the numerals from 1 \textit{yī} \begin{CJK}{UTF8}{gbsn}一\end{CJK} to 10 \begin{CJK}{UTF8}{gbsn}十\end{CJK} \textit{shí}).
Average morphosyntactic complexity is the weighted average of the count of number and arithmetic combinator morphemes ($+,-,*$) across all numerals.
For example, the \textit{numeral} (i.e. the form) of \textit{number} 43 is formed in Mandarin as \begin{CJK}{UTF8}{gbsn}四十三\end{CJK}, \textit{sì shí sān}, ``four [times] ten [plus] three'', $4*10+3$.
This numeral has morphosyntactic complexity of 5, counting the digits $4$ and $3$, the multiplier (base) $10$ and null combinator morphemes $*$ and $+$.

To show that natural recursive systems optimise this trade-off, D\&S compared the numeral systems of natural languages to theoretically possible systems over the number range $1-99$.
They generated these possible recursive numeral systems using Hurford's grammar \citep{Hurford1975TheLT, HURFORD2007773}:
\begin{equation}
\label{eq:hurford-grammar}
\begin{aligned}
Num &\to D \mid Phrase \mid Phrase \pm Num \\
Phrase &\to M \mid Num * M
\end{aligned}
\end{equation}
where $D$ and $M$ are sets of digits and multipliers.

Hurford's grammar allows representing recursive numeral systems succinctly as $(D, M)$ pairs. 
D\&S used a genetic algorithm (cf. Appendix \ref{sec:algorithms}, Algorithm \ref{alg:ga}) to obtain the \textit{Pareto frontier}, the theoretical bound beyond which no possible language can improve on lexicon size without sacrificing average morphosyntactic complexity (or vice-versa), and showed that human recursive numeral systems lie on or close to the frontier.

However, unlike human numeral systems, the numerals in their optimal artificial systems did not consistently employ the same multiplier across numerals (cf. Table \ref{tab:ex_DandS}).
Systems like this are widespread along D\&S's frontier, and were taken to be as ``efficient'' as natural ones, but they are evidently not human-like.
This suggests that optimisation for lexicon size and average morphosyntactic complexity alone may not accurately capture what makes natural recursive numeral systems different from other possible ones.
\begin{table}[ht]
\footnotesize
\centering
\begin{tabular}{crr}
\toprule
Number & Numeral (Mandarin) & Numeral (unnatural) \\
\midrule
18 & $10 + 8$ & $25 - 7$  \\
19 & $10 + 9$ & $25 - 6$  \\
20 & $2 * 10$ & $5 * 4$  \\
21 & $2 * 10 + 1$ & $3 * 7$ \\
22 & $2*10+2$ & $25 - 3$ \\
23 & $2*10+3$ & $25 - 2$ \\
24 & $2*10+4$ & $6 * 4$ \\
25 & $2*10+5$ & $25$ \\
\bottomrule
\end{tabular}
\caption{Numerals in Mandarin (second column) and in one of the optimal artificial systems of D\&S (third column) for the number range $18-25$. Notice that Mandarin is consistently base-10, whereas the unnatural system has alternating bases (i.e., 4, 7 and 25).}
\label{tab:ex_DandS}
\end{table}

Y\&R attempted to solve these issues by introducing ad-hoc constraints over the space of possible numeral systems, resolving the unorderliness issue of D\&S by restricting the numeral system space to be more human-like.
They added an extra rule for suppletives, and four additional constraints, which specify which multipliers to use for which number ranges, the maximum number that can be added or subtracted from ${Phrase}$ within a given number range, as well as exception rules.
They further require that the digits in $D$ run sequentially from $1$ to a number between $1-20$. They also only allowed for the smallest multiplier to be used to construct numbers smaller than or equal to the next largest multplier in $M$ (Other examples can be found in \citealp{yang2025re}).

Crucially, Y\&R still maintain that lexicon size and morphosyntactic complexity are the relevant measures.
In this sense, D\&S's optimal languages are still considered as efficient as Y\&R ones, even though they cannot be generated under the grammar constraints used by Y\&R.

\section{(Ir)regularity and processing complexity}\label{sec:irregularityandproccomp}

We find the resolution proposed by Y\&R unsatisfying, since the source of these additional constraints itself stands in need of explanation; ideally, these characteristics could be accounted for as byproduct of other better-motivated pressures on natural recursive numeral systems. Importantly,  neither of the existing measures captures a crucial aspect of human systems, namely \textit{regularity}.
An ``unorderly'' system, such as one with alternating multipliers, is intuitively not human-like precisely because the numerals are irregular, and Y\&R constraints yield more human-like systems precisely because they are explicit constraints enforcing regularity.
We propose that recursive numeral systems are better thought of as optimising the trade-off between irregularity and processing complexity, and provide measures of each below.

\subsection{Irregularity}
One way to measure regularity is in terms of \textit{compressibility}.
Regular systems are highly compressible, such that one could write a small set of general rules for the construction of numerals.
For instance, recursive numerals in Karo Batak \citep{woollamsGrammarKaroBatak1996} are extremely regular.
For numerals from 1 to 99, it only requires the rules $Num \rightarrow{} D$ (for 1-9), and $Num \rightarrow{} D*10+D$ (for 10-99).

\begin{figure*}
    \centering
    \includegraphics[width=0.7\textwidth]{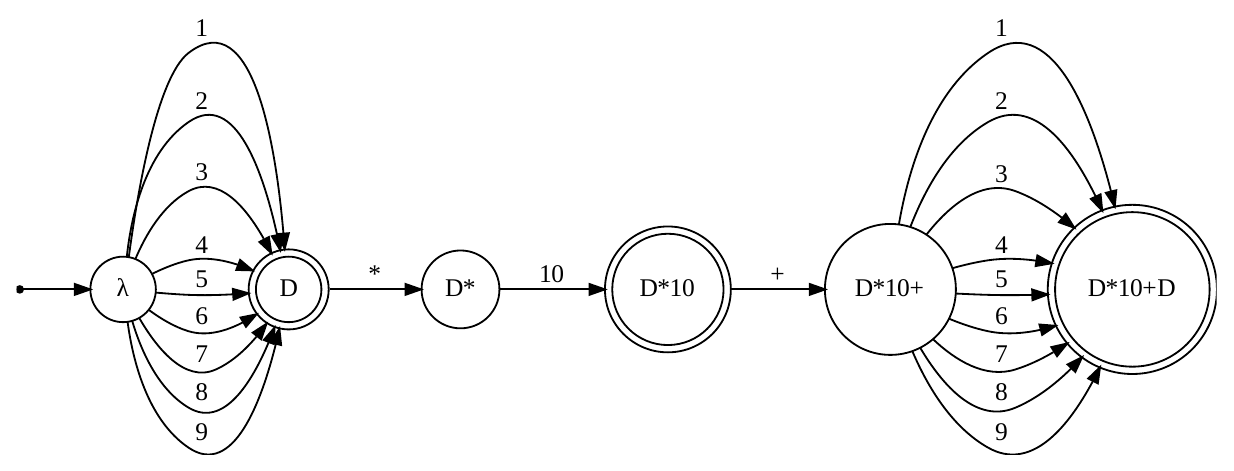}
    \caption{The minimal partial DFA representing Karo Batak (highly regular) numerals in the range 1 to 99.
    Circles are states. Double circles are accepting states. Arrows are transitions.
    $\lambda{}$ denotes the initial state.
    A given numeral is parsed or generated by traversing the automaton and emitting the symbols (morphemes) associated with the transitions.
    Accepting states mark valid potential termination points of the parsing or generation.
    Note that, here, we name each state in accordance with the rules to generate (partial) numerals when the state is reached.
    For example, the number 10 in Karo Batak, expressed by $1*10$, corresponds to the path from $\lambda$ through the transition for $1$ to the state $D$, then the transition for $*$ to the state $D*$, then the transition for $10$, ending at the state $D*10$.
    }
    \label{fig:simpledfa}
\end{figure*}

English numerals are slightly less regular, i.e. featuring suppletive numerals ($11$, $12$) that require separate rules.
We can thus measure the (ir)regularity of a system by finding the length of the most general/shortest grammar for numerals.
However, as inferring the shortest recursive grammar is challenging \citep{charikarSmallestGrammarProblem2005}, we simplify the problem by restricting ourselves to the number range 1-99, following D\&S and Y\&R, and treating the system as a finite language.
We inferred a minimal partial deterministic finite automaton (DFA) that generates all and only valid numerals to represent a numeral grammar using \citeauthor{mihovFiniteStateTechniquesAutomata2019}'s (\citeyear{mihovFiniteStateTechniquesAutomata2019}) algorithm.\footnote{We use the Python package \texttt{automata-lib} (\texttt{\url{https://pypi.org/project/automata-lib/}}) for inference.}
Intuitively, a regular system has a lot of repeated parts, so the number of states and transitions of the DFA will be small (cf. Figure \ref{fig:simpledfa}).

Our measure of irregularity is thus the representational complexity of the numeral system, i.e., the size of the partial DFA.
Adapting the scheme of \citet{brighton2003simplicity}, the irregularity $L(G)$ of a recursive numeral grammar $G$, in bits, is:
\begin{equation}\label{eq:lG}
    \underbrace{|Z|\bigl(2\cdot{}log_2|S| + log_2|\Sigma|\bigr)}_{\text{(A)}} + \underbrace{log_2|S|}_{\text{(B)}} + \underbrace{|S|}_{\text{(C)}}
\end{equation}
where $Z$ is the set of transitions, $S$ is the set of states, and $\Sigma$ is the set of symbols (i.e., morphemes).  
(A) gives the number of bits required to encode the transition table: Each transition requires $2 \cdot{} log_2|S|$ bits to identify the states at the start and end of the transition, and $log_2|\Sigma|$ to identify the transition symbol (morpheme).
(B) and (C) are the numbers of bits for identifying the single initial state and accepting states (one bit for each state, indicating whether it is an accepting state or not), respectively.

\subsection{Processing complexity}
The DFA can be thought of as a description language used to encode the data (a set of numerals). 
The more complex the description language (the larger the DFA), the easier it is to generate or parse a numeral.\footnote{An anonymous reviewer asked if the two measures are actually independent (and hence could trade off each other). Impressionistically, the distribution of the baseline languages in Figure \ref{fig:randomsamplecomp} suggests that they are. 
Following the reviewer's advice, we have also provided empirical correlations between regularity, processing complexity and average morphosyntactic complexity in Appendix \ref{sec:correlation}.}
Drawing on the Minimum Description Length principle \citep{li2008introduction}, \citet{brighton2003simplicity} shows that this quantity can be calculated from the average information required to determine the unique parsing path through the DFA for a given form-meaning pairing.
In our case, this pairing simply corresponds to a numeral (the form) as one morpheme refers to only one meaning (e.g., \textit{one} in \textit{twenty one} refers only to `1').
For instance, in the DFA shown in Figure \ref{fig:simpledfa}, parsing $2*10$ (20) requires going through 4 transitions connecting the states $\lambda{}$ (initial state), $D$, $D*$ and $D*10$, and one would need $\underbrace{log_2(9)}_{\lambda}+\underbrace{log_2(1)+1}_{D}+\underbrace{log_2(1)}_{D*}+\underbrace{log_2(1)+1}_{D*10}$ bits to encode or generate this numeral as a path through the DFA, with each accepting state requiring an extra bit to encode the accepting status.

Averaging over these bits for all numbers gives the \textit{processing complexity} of generating or parsing a set of numerals using a given DFA.
Again adapting \citeauthor{brighton2003simplicity}'s (\citeyear{brighton2003simplicity}) scheme,
the processing complexity $L(N\mid{}G)$ of recursive numeral grammar $G$ for a set of numbers $N$ is given by:
\begin{equation}
  \sum_{i=1}^{|N|} \Bigl\{ P(n_i) \cdot{} \sum_{j=1}^{|S_i|}\bigl(log_2|Z_{ij}| + F(s_{ij}) \bigr)\Bigr\}
\end{equation}
where $P(n_i)$ is the prior/weight for $n_i \in N$, $S_i$ is the list of states through which the parsing proceeds for the numeral for $n_i$, $Z_{ij}$ is the set of outward transitions from the $j$th state in the parsing of the numeral for $n_i$, and $F$ is an indicator function evaluating to 1 if $s_{ij} \in S_i$ is an accepting state.
D\&S and Y\&R assume that the prior is $P(n_i)
\propto n_i^{-2}$, taken to reflect the communicative need of numbers \citep{dehaene1992cross,piantadosi2016rational}, i.e. there is high communicative need / frequent use of small numbers, and the need to express higher numbers declines rapidly.
We follow this assumption in Section \ref{sec:applying}
but challenge it in Section \ref{sec:prior}.

\section{Applying the measures}\label{sec:applying}

We now apply our complexity measures to natural recursive numeral systems and compare them to other theoretically possible ones.

\subsection{Natural vs. baseline numeral systems}\label{sec:langcomparison}
We first ask if natural recursive numerals are particularly regular and of low processing complexity, relative to the entire space of possible systems. 
We compute our measures on two datasets: (1) the recursive numeral systems from 128 natural languages annotated at the morpheme level from D\&S,\footnote{\texttt{\url{https://osf.io/cmuq4}}} and (2) a sample of 10,000 baseline artificial systems generated with Hurford's grammar.
To obtain the baseline sample, we adapt the implementation of the genetic algorithm in D\&S.
When constructing artificial numeral systems, in cases where a grammar allows a number to be expressed as several possible numerals, D\&S always chose the shortest numeral for each number (e.g. given $D = \{1,2\}, M=\{5, 10, 12\},$ the possible numerals for 12 include the shortest $12$, along with $1*12$, $10+2$, $1*10+2$ and $2*5+2$).
This is partly the source of unorderliness: favouring the shortest possible numeral will result in a system that does not reuse structures in smaller numerals in bigger ones (e.g., using $12$ instead of $10+2$ when the numeral for 11 is $10+1$).
To alleviate this, we allow numerals of any length and any combination of digits, multipliers and combinators ($+, *, -$).
For computational tractability, however, we limit the max search depth to 5, meaning that a numeral has at most 5 number morphemes and 4 arithmetic combinators, and use only digits and multipliers that occur in at least one natural language.
We draw languages in 100 batches of 100 language types.
We determine the type for each batch as follows.
We randomly select 3 to 12 digits and 1 to 3 multipliers.
We always allow $+$ and $*$ as combinators, with a 20\% chance of also having the subtraction combinator $-$, which is less common among the world's languages \citep{comrie2022arithmetic}.
We resample if the resulting set of digits, multipliers and combinators cannot express all numbers between 1 and 99.
Otherwise, we use the selected digits, multipliers and combinators to generate 100 sets of numerals, each covering the numbers 1-99, where those samples can differ in how a given number is expressed (cf. the multiple possible numerals for 12 above).

\begin{figure}[hbt]
    \centering
\includegraphics[width=0.9\linewidth]{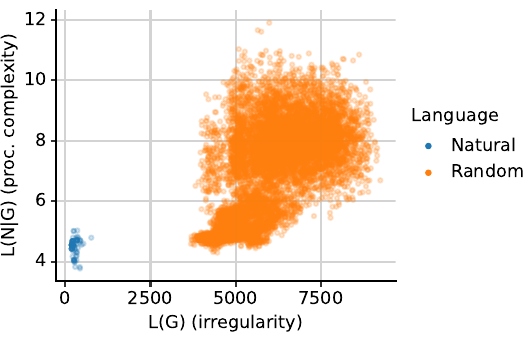}
    \caption{Irregularity (\textit{x}-axis) and processing complexity (\textit{y}-axis) of natural languages (blue) and 10,000 randomly generated baseline artificial languages (orange).}
    \label{fig:randomsamplecomp}
\end{figure}

Figure \ref{fig:randomsamplecomp} plots irregularity against processing complexity for natural systems and these baseline systems. 
As expected, natural systems exhibit significantly more regularity and less processing complexity than the baselines.
In fact, none of the sampled systems are simultaneously more regular and less complex than natural systems.

Next, we ask whether our measures successfully capture the key properties of natural systems that distinguish them from other possible systems, rather than simply recapitulating the previous measures.
We compute our measures on two more datasets: (1) optimal artificial systems evolved through the genetic algorithm from D\&S, and (2) optimal artificial systems evolved through the genetic algorithm with the constraints from Y\&R.
\begin{figure}[ht]
    \centering
\includegraphics[width=0.9\linewidth]{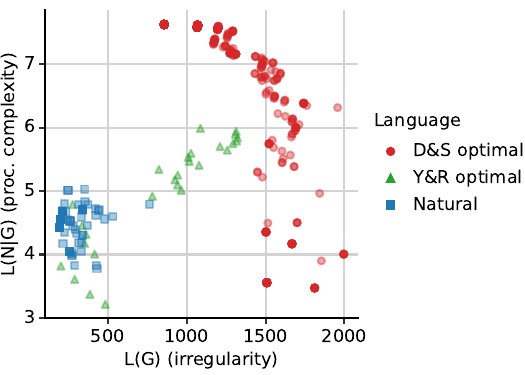}
    \caption{Irregularity (\textit{x}-axis) and processing complexity (\textit{y}-axis) of natural systems (blue, square) and optimal systems in D\&S (red, circle) and Y\&R (green, triangle)}
    \label{fig:optimalcomp}
\end{figure} 

Figure \ref{fig:optimalcomp} shows that the different systems exhibit varying degrees of regularity and processing complexity, with natural systems typically being more efficient than the optimal systems according to the metrics and grammars used by D\&R and Y\&R.
Interestingly, among Y\&R systems, which already have some degrees of regularity built in, there are some that are clearly less efficient than natural systems (the middle cluster), and crucially a few that appear to be \textit{more} efficient than natural systems (bottom left corner).
Upon inspection, the less efficient systems use a large set of digits and large mutlipliers, e.g., digits up to 20 and $M$ = \{21, 42, 56, 71, 85\}, which results in higher irregularity (cf. $|\Sigma|$ in (\ref{eq:lG})), and processing complexity due to more transitions when constructing numerals in higher ranges.
The four highly efficient Y\&R numeral systems are those with digits and multiplier in the range 1 to 5.
In other words, these have an extremely small lexicon size with digits in the smallest ranges.
This allows them to reuse smaller numerals to build up larger ones, thereby increasing regularity at little cost of processing complexity  (e.g., 5 expressed as $2*2+1$; 6 as $(2+1)*2$;  7 as $(2+1)*2+1$; 8 as $2*2*2$; 9 as $2*2*2+1$; 10 as $(2*2+1)*2$).
Larger numerals in these systems end up with high processing complexity (e.g., 46 is expressed as $(((2*2+1)*2+1)*2+1)*2$), but contribute little to the average processing complexity because the power law prior places little weight on larger numbers.

\subsection{Controlling for digits, multipliers, combinators, and numeral lengths}\label{sec:control}
Most of the D\&S and Y\&R optimal systems in Figure \ref{fig:optimalcomp} do not necessarily feature the same \textit{combinations} of digits and multipliers as natural languages.
One might wonder, then, whether the apparent regularity and processing simplicity of natural recursive numeral systems could still result from the optimisation for the trade-off between lexicon size and average morphosyntactic complexity, in conjunction with the specific $(D, M)$s that natural systems employ.
Indeed, the Y\&R inefficient systems are the ones with unnatural (D, M)s, e.g., $M=\{21,42,56,71,85\}$.
One response is to say that the choice of human-like $(D, M)$ pairs themselves would still demand an explanation, which regularity and processing simplicity could provide.
Here, however, we seek an even stronger piece of evidence.
We aim to show that natural recursive numeral systems are (near-)optimally regular and of low processing complexity \textit{even} among their \textit{local neighbourhood}.
Specifically, each natural numeral system's local neighbourhood is the set of possible systems that share its digits $D$, multipliers $M$, arithmetic combinators $C$ and the lengths of numerals $L_{Num}$.
For example, every system in Drehu's local neighbourhood will have numerals that make use of $D=\{1,2,3,4\}$, $M=\{5, 10, 15,20\}$, $C=\{*,+\}$ and match Drehu's numerals in length.
To illustrate, Drehu $24$ is expressed as $1*20+4$, so its alternative numerals are $\{4*5+4, 5+15+4, 15+5+4, 10+10+4\}$.
Note that, because of this, every system in a local neighbourhood will also share the same lexicon size and average morphosyntactic complexity, effectively controlling for these measures.

Since exhaustively constructing the local neighbourhoods is computationally prohibitive, we implement a greedy algorithm to approximate only the most (and least) efficient numeral systems.
For each natural system, we consider adding alternative numerals starting with the largest numbers and working towards the lower range of numbers; from the set of such perturbations, we weed out all systems apart from the ones on the Pareto frontier, and repeat the process until we obtain full systems covering the number range 1--99. 
The full procedure is explained and given as pseudocode alongside our parameter choices in Algorithm \ref{alg:control}, Appendix \ref{sec:appendix_control}. 
The expectation is that natural systems should consistently be as efficient or more efficient than the local frontier from this algorithm.

\begin{figure*}[htb]
    \centering
    \includegraphics[width=\textwidth]{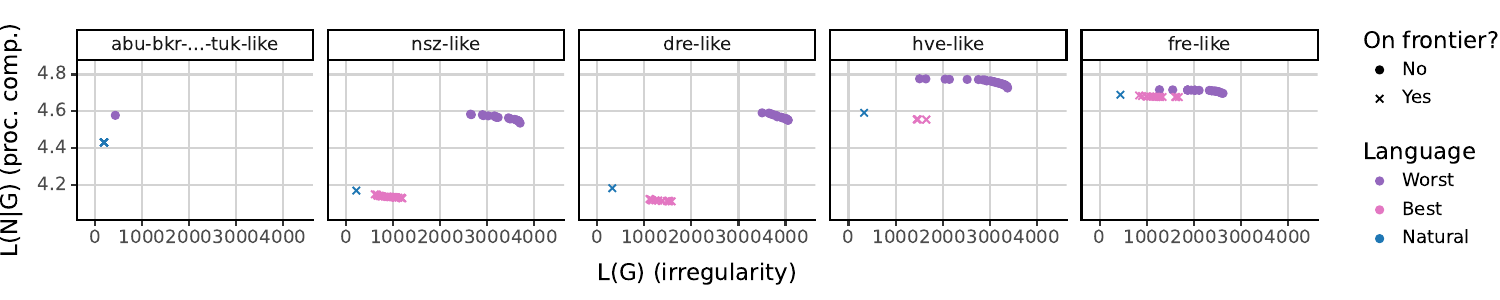}
    \caption{Irregularity (\textit{x}-axis) and processing complexity (\textit{y}-axis) of the most (purple) and least (pink) efficient natural-$(D, M, C, L_{Num})$-matched artificial systems, as well as their corresponding natural systems (blue). 
    Natural systems are consistently on the approximately local frontier (cross).
    Leftmost panel shows a case where the space is very small with one best--occupied by natural systems like Karo Batak--and one worst system. 
    Subsequent panels (Sierra Nahuatl-, Drehu-, Huave- and French-like systems) show cases where there is room for variation, though $L(N\mid G)$ varies to different degrees depending on exact $(D, M, C, L_{Num})$. French-like systems exhibit little variation in $L(N\mid G)$, whereas Nahuatl-like systems exhibit more.}
    \label{fig:localspace}
\end{figure*}

We obtain the frontiers for 37 unique local neighbourhoods inferred from 127 out of the 128 numeral systems of natural languages.\footnote{We exclude Kayah Li's neighbourhood due to limited computational resources. cf. Section \ref{sec:appendix_control}, Appendix \ref{sec:algorithms}.}
We find that, while the spaces are (unsurprisingly) small, 121 out of 127 natural systems are either as efficient or more efficient than their local frontiers. 
While for some of these, the spaces are so small that there are only a couple of possible systems (cf. the space for Karo Batak-like systems in the leftmost panel of Figure \ref{fig:localspace}, which has only two possible systems under our constraints: best and worst), there are larger spaces with clearly more room for variation, and we  found that natural systems are consistently on the local frontiers (Panels 2-4 in Figure \ref{fig:localspace} illustrate typical results; consult the full plots on \href{https://github.com/ponrawee/regularity-processing-complexity-recursive-numerals}{our GitHub repository}).
Overall, this suggests that optimisation for lexicon size, average morphosyntactic complexity, along with $(D, M, C, L_{Num})$ combinations, do not fully explain the degree of regularity and processing complexity that natural systems exhibit.

\section{The role of the prior}\label{sec:prior}
 Previous work has argued that recursive numeral systems are subjected to a power law prior distribution $P(n)
 \propto n^{-2}$, which reflects communicative need over numbers \citep{dehaene1992cross,xu2020numeral, denic2024recursive,yang2025re}.
 This presents a very strong bias towards the smallest numbers, as when calculating either average morphosyntactic or processing complexity most numerals will have minimal influence. More specifically, when considering the numeral range $1-99$, more than $75\%$ of the probability mass is placed on numbers $1$ and $2$ alone, and more than $90\%$ is placed on numbers $1-6$. 
 This is why, for example, the optimal system in Section \ref{sec:langcomparison} with incredibly complex large numerals (e.g. a system where the number 46 expressed as the numeral $(((2*2+1)*2+1)*2+1)*2$) is considered easier process than natural systems: these numerals contribute negligibly to the final result when considering the power law prior. 
 
However, the purpose of recursive numeral systems is to be able to precisely express the higher-range numbers that are heavily downweighted by the power law prior. 
 If the larger numbers' influence was so negligible in reality, one would expect  recursive numeral systems to be less widespread, as exact-restricted systems like Kayardild, which has a unique word for numbers 1--4 and a single word covering all numbers greater than $4$ (\textit{muthaa}), would seem as effective and much easier to learn.

\begin{figure}[tb]
    \centering
\includegraphics[width=0.9\linewidth]{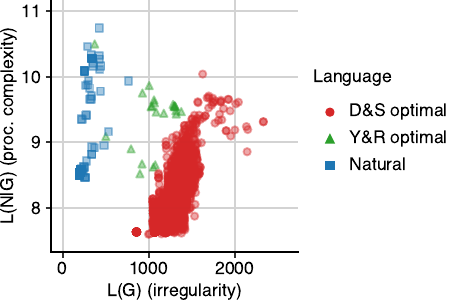}
    \caption{Irregularity (\textit{x}-axis) and unweighted processing complexity (\textit{y}-axis) of natural systems (blue, square) and optimal systems in D\&S (red, circle) and Y\&R (green, triangle).}
    \label{fig:optimaluniform}
\end{figure}

\begin{figure}[htb]
    \centering
\includegraphics[width=0.9\linewidth]{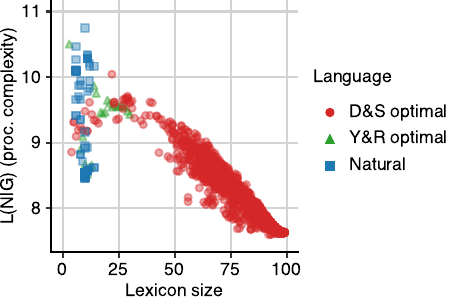}
    \caption{Lexicon size  (\textit{x}-axis) and unweighted processing complexity (\textit{y}-axis) of natural systems (blue, square) and optimal systems in D\&S (red, circle) and Y\&R (green, triangle).}
    \label{fig:uniflexicon}
\end{figure}

Therefore, there might be reason to believe that these sorts of systems are not influenced by a need distribution which emphasises small numbers so heavily.
Instead, a uniform prior over numbers may better reflect the function of recursive numeral systems.
To investigate the role of the prior, we obtain systems that are optimal in terms of lexicon size and average morphosyntactic complexity, in the same vein as D\&S and Y\&R, except under a uniform prior.
We then compare these new optimal systems to human systems in terms of irregularity and (uniformly weighted) processing complexity.
As Figure \ref{fig:optimaluniform} shows,\footnote{We exclude two outlier Y\&R/D\&S systems with extremely high complexities from Figure \ref{fig:optimaluniform} and \ref{fig:uniflexicon}.} while natural systems still exhibit high regularity, the optimal artificial systems lie much further right, with Y\&R systems being slightly more regular than D\&S, but still far from human systems. 
Notably, the four Y\&R systems in Section \ref{sec:langcomparison} are no longer considered more efficient than the human ones under the uniform prior.

Surprisingly, many D\&S optimal systems end up being more efficient than human systems in terms of processing complexity under this prior. 
This is likely because, unlike human and Y\&R systems, D\&S do not constrain the lexicon size and always choose the shortest numerals possible.
Thus, when the lexicon size is large, the processing complexity is low (Figure \ref{fig:uniflexicon}).

For example, a D\&S system with all number morphemes from 1 to 99 will express every number with a single morpheme, and will have a single accepting state and $99$ transitions to it, resulting in $L(N
\mid G) = log_2(99)+1 \simeq 7.63$ bits per number.

On the other hand, to parse a numeral like $96$ in a highly regular system like Karo Batak (Figure \ref{fig:simpledfa}), we would need $\underbrace{log_2(9)}_{\lambda}+\underbrace{log_2(1)+1}_{D}+\underbrace{log_2(1)}_{D*}+\underbrace{log_2(1)+1}_{D*10} +$ $+ \underbrace{log_2(1)}_{D*10+} + \underbrace{log_2(9)+1}_{D*10+D} \simeq 9.34 > 7.63$ bits. 
While the possible transitions are fewer, Karo Batak is harder to process than a 99-word system due to the extra bits required by the larger number of accepting states. 

Regardless of the shape of the correct prior over numbers, our results show how natural systems are much more efficient in both measures (Figure \ref{fig:optimalcomp}) or at least in regularity (Figure \ref{fig:optimaluniform})  than the optimal artificial systems obtained by both D\&S and Y\&R when optimizing for lexicon size and average morphosyntactic complexity.

\section{Discussion}\label{sec:discussion}
We have argued that regularity is a major pressure that shapes linguistic structures, and should be included in accounts of how efficient communication explains constrained variation.
In particular, we have argued that prior literature on the efficiency of \textit{recursive numeral systems} \citep{denic2024recursive,yang2025re} has not satisfactorily accounted for regularity in human-like systems, as well as processing complexity, which trades off with regularity.
In this paper, we have shown that human recursive numeral systems are more regular and easier to process than other theoretically possible systems, even when controlling for the set of number morphemes and numeral lengths.
We are skeptical that the relevant prior is the power law communicative need distribution \citep{dehaene1992cross,piantadosi2016rational} as assumed in prior work \citep{xu2020numeral,denic2024recursive,yang2025re}.
We show that even if the relevant prior was uniform, our measures would still clearly separate human-like and other possible recursive numeral systems. 

One issue that we have not considered here is the role of cognitive cost of encoding certain meanings or relations.
For example, systems using highly unnatural numerals like $(((2*2+1)*2+1)*2+1)*2$ (Section \ref{sec:langcomparison}) are currently considered efficient.
We have shown that this might be because of the power law prior assumption (Section \ref{sec:prior}), but the absence of this kind of numeral could also be due to the high cost of computation involved in the multiply nested multiplications, which are harder to learn in numeral systems than operations like additions \citep{holt2025learning,silvi2025le}, while our measure of processing complexity assumes that all transition symbols are equally costly.
Similarly, expressing numerals with multiple nested parentheses would also introduce ambiguity in how they are interpreted (e.g., $2*10+5$ could be either $(2*10)+5$ or $2*(10+5)$).

Another issue concerns the granularity of the annotated data from D\&S. 
We notice that the annotation was not sensitive to the order of morphemes in each numeral, and they would always annotate the numerals according to Hurford's left-headed grammar.
For example, English \textit{-teen}s were annotated as $10+D$ instead of the real order $D+10$ (e.g., four-teen should be $4+10$).
The same is true for multiplications; the product of $D$ and $M$ is annotated as $D*M$.
For example, \textit{\` iri īse} `50' in Igbo is $10*5$ \citep{emenanjo1987elements} but was annotated $5*10$. This presents no problem for their measures, which are not sensitive to morpheme order. However, our measures are, and this annotation results in overestimated regularity, since, for instance, the orders of $M$ and $D$ in natural languages often change depending on number ranges, reducing regularity. 
We have begun the process of reannotating their data, in order to re-run the analyses here taking this order variation into account.
Early results suggest that the order-sensitive reannotation does not change our qualitative results, at least with regard to the efficiency of recursive numeral systems relative to other theoretically possible, including D\&S- and Y\&R-optimal systems.
This is likely because natural systems maintain a high degree of regularity in order over large number ranges (e.g., only use $M+D$ over the teens range, as in English).

Unlike most work in the efficient communication literature, which only considers the \textit{mapping} between form and meaning (cf. \citealp{kemp2018semantic}), work on recursive numeral systems like ours, D\&S and Y\&R focuses on the \textit{forms} themselves.
Our work specifically examines the \textit{relationship} among forms, i.e., the regularity of the system, and shows that it is a vital property of efficient, human-like linguistic systems.
This suggests, for example, that simplicity might be underestimated in work that treats compositional signals as equally complex as unanalysable ones (e.g., in grammatical categories like person systems; \citealp{Zaslavsky2021person}).
Moreover, our re-examination of D\&S and Y\&R also suggests that it is not sufficient simply to find a set of measures under which natural numeral systems could be shown to be optimal.
To actually pinpoint why certain systems are human-like, one would want to identify the properties, like regularity and processing complexity, which explain the natural preferences of human-like systems over other possible ones, and which have plausible cognitive bases.

Our approach bears some relationships to and has the potential to improve upon other approaches to constrained cross-linguistic variation.
For example, much previous work has suggested that some systems are more widespread than others thanks to their learnability \citep{gentner2009some,culbertson2012learning,STEINERTTHRELKELD2020104076,prasertsom2026domain}.
As high regularity and low processing complexity are two aspects of simplicity, they also make some types of numeral systems easier to learn, and thus more likely to be typologically widespread \citep[cf.][]{culbertson2016simplicity}.
Relatedly, our idea of relating complexity in terms of compressibility (regularity) to constrained variation is not new.
For instance, \citet{van2023quantifiers} show that the monotonicity of natural language quantifiers stem from the fact that it yields more compressibility (shorter description length). 
In fact, experimental research in language evolution explicitly demonstrates that pressures from learning drive compressibility \citep[e.g.][]{kirby2015compression,carrSimplicityInformativenessSemantic2020}.
A recent approach makes use of the Information Bottleneck (IB) principle \citep[e.g.][]{zaslavsky2018efficient}, which provides a mathematically precise way to characterise the simplicity-informativeness trade-off in terms of mutual information, but has so far not taken into account the systematicity of forms.
An interesting direction would be to information-theoretically formalise regularity and processing complexity under the IB framework.

One clear extension of our work here is an investigation into whether natural recursive numeral systems are on the Pareto frontier of the irregularity/processing complexity trade-off. 
This is particularly challenging, because the space of all possible recursive numeral systems is extremely large, even when we only consider Hurford-grammar-conforming systems and limit the maximum search depth. Unlike D\&S and Y\&R, we would also like to avoid constraining the space by simply favouring shortest numerals or adding ad-hoc constraints.
One potential solution is to adopt Y\&R's enhanced grammars as representations while dropping their ad-hoc implementational constraints, allowing us to consider a much larger solution space while still keeping compact recursive numeral representations. 
Another complementary path would be to provide a mechanistic explanation of how these systems have
been shaped over time via general learning mechanisms such as reinforcement learning, similar to \citet{carlsson2021learning} and \citet{silvi2025le}, and also via behavioural experiments, both of which could investigate the effects of regularity on learnability as suggested above.

Finally, a more general study that fully incorporates form-meaning mappings would yield a unifying model for numeral systems, spanning from recursive numeral systems, which have one-to-one form-meaning mappings, to exact-restricted and approximate ones, which have one-to-many mappings.
In fact, a general approach to measuring form-meaning systematicity would enable a fuller exploration of efficiency in any domain whose linguistic signals range from wholly unanalysable to fully compositional.
It would also enable us to explore broader questions on linguistic complexity, such as whether different systems of the same language trade off in complexity \citep[e.g. ][]{bentz2023complexity}. 
We hope to continue the work in these directions in the future.

\section{Acknowledgments}
The authors would like to thank anonymous reviewers for constructive feedbacks and suggestions.
P.P. is supported by Anandamahidol Foundation, Thailand. 
A.S. is supported by the Swedish Research Council.

\section{Limitations}

First, we have limited computational resources, so we were not able to use a higher search depth to explore even larger spaces of possible numeral systems.
However, as Figure \ref{fig:randomsamplecomp} shows, the space we explore is already larger than that considered in previous work (D\&S and Y\&R) combined.

The second limitation concerns the format of the data. As we pointed out in the discussion, the D\&S annotation always employs the same order (i.e., consistent with Hurford's left-headed grammar), which may inflate estimated regularity under our measure. While preliminary results suggest that this does not change the key findings, in future work we will analyse the newly annotated data in more detail.

The next limitation also concerns the data.  D\&S used the language sample from the World's Atlas of Language Structures (WALS) \citep{wals-s1}.
While, as they pointed out, WALS was constructed with a focus on language diversity, we found that the sample comprises many languages from large language families such as Indo-European and Austronesian.
This could be taken as ``representative'' in so far as the sample is roughly proportional to the number of languages across families, but a more appropriate sample might be one with more balance across large and small families, and which is controlled for geographical effects \citep{dryer1989large,laddCorrelationalStudiesTypological2015}. 

Finally, while our work is focused on comparing the efficiency of natural and other theoretically possible recursive numeral systems, including the previously ``optimal'' systems obtained in previous work, one might still wonder how efficient natural numeral systems actually are relative to the theoretical optima. 
D\&S and Y\&R show that natural recursive numeral systems lie close to their Pareto frontier, and, as noted in the discussion, we hope to do the same in future work.

\bibliography{anthology,custom}

@book{chomsky2002syntactic,
  title={Syntactic structures},
  author={Chomsky, Noam},
  year={2002},
  publisher={Walter de Gruyter}
}

@incollection{gentner2009some,
  title={Why some spatial semantic categories are harder to learn than others: The typological prevalence hypothesis},
  author={Gentner, Dedre and Bowerman, Melissa},
  booktitle={Crosslinguistic approaches to the psychology of language: Research in the tradition of Dan Isaac Slobin},
  pages={465--480},
  year={2009},
  publisher={Psychology Press}
}

@book{jackendoff2002foundations,
  title={Foundations of language},
  author={Jackendoff, Ray},
  year={2002},
  publisher={Oxford University Press}
}

@article{denic2024recursive,
  title={Recursive Numeral Systems Optimize the Trade-off Between Lexicon Size and Average Morphosyntactic Complexity},
  author={Deni{\'c}, Milica and Szymanik, Jakub},
  journal={Cognitive Science},
  volume={48},
  number={3},
  pages={e13424},
  year={2024},
  publisher={Wiley Online Library},
doi={https://doi.org/10.1111/cogs.13424}
}

@article{denicIndefinitePronounsOptimize2022a,
  title = {Indefinite {{Pronouns Optimize}} the {{Simplicity}}/{{Informativeness Trade}}‐{{Off}}},
  author = {Denić, Milica and Steinert‐Threlkeld, Shane and Szymanik, Jakub},
  date = {2022-05},
  journal = {Cognitive Science},
  shortjournal = {Cognitive Science},
  volume = {46},
  number = {5},
year={2022},
  pages = {e13142},
  issn = {0364-0213, 1551-6709},
  doi = {10.1111/cogs.13142},
  url = {https://onlinelibrary.wiley.com/doi/10.1111/cogs.13142},
  urldate = {2025-10-06},
  langid = {english},
}

@article{piantadosi2016rational,
  title = {A Rational Analysis of the Approximate Number System},
  author = {Piantadosi, Steven T.},
  year = {2016},
  month = jun,
  journal = {Psychonomic Bulletin \& Review},
  volume = {23},
  number = {3},
  pages = {877--886},
  issn = {1069-9384, 1531-5320},
  doi = {10.3758/s13423-015-0963-8},
  urldate = {2023-06-12},
  langid = {english},
}

@article{chen2023information,
  title={An information-theoretic approach to the typology of spatial demonstratives},
  author={Chen, Sihan and Futrell, Richard and Mahowald, Kyle},
  journal={Cognition},
  volume={240},
  pages={105505},
  year={2023},
  publisher={Elsevier}
}

@article{kemp2018semantic,
  title={Semantic typology and efficient communication},
url={https://doi.org/10.1146/annurev-linguistics-011817-045406},
  author={Kemp, Charles and Xu, Yang and Regier, Terry},
  journal={Annual Review of Linguistics},
  volume={4},
  number={1},
  pages={109--128},
  year={2018},
  publisher={Annual Reviews}
}

@inproceedings{Zaslavsky2021person,
	Author = {Zaslavsky, Noga and Maldonado, Mora and Culbertson, Jennifer},
	Title = {Let's talk (efficiently) about us: {P}erson systems achieve near-optimal compression},
	Booktitle = {Proceedings of the 43rd Annual Meeting of the Cognitive Science Society},
	Year = {2021},
url={https://escholarship.org/uc/item/2sj4t8m3}
}

@article{dryer1989large,
  title={Large linguistic areas and language sampling},
url={https://doi.org/10.1075/sl.13.2.03dry},
  author={Dryer, Matthew S},
  journal={Studies in Language. International Journal sponsored by the Foundation “Foundations of Language”},
  volume={13},
  number={2},
  pages={257--292},
  year={1989},
  publisher={John Benjamins}
}

@article{laddCorrelationalStudiesTypological2015,
  title = {Correlational {{Studies}} in {{Typological}} and {{Historical Linguistics}}},
  author = {Ladd, D. Robert and Roberts, Se{\'a}n G. and Dediu, Dan},
  year = {2015},
  journal = {Annual Review of Linguistics},
  volume = {1},
  number = {1},
  pages = {221--241},
  issn = {2333-9683, 2333-9691},
  doi = {10.1146/annurev-linguist-030514-124819},
  urldate = {2025-10-06},
  langid = {english},
}

@book{emenanjo1987elements,
  title={Elements of modern Igbo grammar: A descriptive approach},
  author={Emenanjo, Emmanuel Nolue},
  year={1987},
  publisher={Oxford University Press}
}

@article{gibson2019efficiency,
  title={How efficiency shapes human language},
  author={Gibson, Edward and Futrell, Richard and Piantadosi, Steven P and Dautriche, Isabelle and Mahowald, Kyle and Bergen, Leon and Levy, Roger},
  journal={Trends in cognitive sciences},
  volume={23},
  number={5},
  pages={389--407},
  year={2019},
  publisher={Elsevier},
url={https://doi.org/10.1016/j.tics.2019.09.005}
}

@article{mollica2021forms,
  title={The forms and meanings of grammatical markers support efficient communication},
  author={Mollica, Francis and Bacon, Geoff and Zaslavsky, Noga and Xu, Yang and Regier, Terry and Kemp, Charles},
  journal={Proceedings of the National Academy of Sciences},
  volume={118},
  number={49},
  pages={e2025993118},
  year={2021},
  publisher={National Academy of Sciences},
    url={https://doi.org/10.1073/pnas.2025993118}
}

@article{kemp2012kinship,
  title={Kinship categories across languages reflect general communicative principles},
  author={Kemp, Charles and Regier, Terry},
  journal={Science},
  volume={336},
  number={6084},
  pages={1049--1054},
  year={2012},
  publisher={American Association for the Advancement of Science},
url={https://doi.org/10.1126/science.1218811}
}

@inproceedings{silvi2025le,
  title={Learning Efficient Recursive Numeral Systems via Reinforcement Learning},
  author={Silvi, Andrea and Thomas, Jonathan and Carlsson, Emil and Dubhashi, Devdatt and Johansson, Moa},
  booktitle={Proceedings of the Annual Meeting of the Cognitive Science Society},
  volume={47},
  year={2025},
url={https://escholarship.org/uc/item/3cc5053z}
}

@inproceedings{yang2025re,
  title={Re-examining the tradeoff between lexicon size and average morphosyntactic complexity in recursive numeral systems},
  author={Yang, David and Regier, Terry},
  booktitle={Proceedings of the Annual Meeting of the Cognitive Science Society},
  volume={47},
  year={2025},
url={https://escholarship.org/uc/item/5k8646v4}
}

@book{li2008introduction,
  title={An introduction to Kolmogorov complexity and its applications},
  author={Li, Ming and Vit{\'a}nyi, Paul and others},
  volume={3},
  year={2008},
  publisher={Springer}
}

@book{woollamsGrammarKaroBatak1996,
  title = {A Grammar of {{Karo Batak}}, {{Sumatra}}},
  author = {Woollams, Geoff},
 year = {1996},
  series = {Pacific Linguistics},
  number = {130},
  publisher = {Dept. of Linguistics, Research School of Pacific Studies, Australian National University},
  location = {Canberra},
  isbn = {978-0-85883-432-3},
  pagetotal = {351}
}

@article{holt2025learning,
  title={Learning a Novel Number System: The Role of Compositional Rules and Counting Procedures},
  author={Holt, Sebastian and Barner, David},
  journal={Cognitive Science},
  volume={49},
  number={6},
  pages={e70071},
  year={2025},
url={https://doi.org/10.1111/cogs.70071},
  publisher={Wiley Online Library}
}

@article{comrie2022arithmetic,
  title={The arithmetic of natural language: Toward a typology of numeral systems},
  author={Comrie, Bernard},
  journal={Macrolinguistics},
  volume={10},
  number={1},
  pages={1--35},
  year={2022}
}

@article{charikarSmallestGrammarProblem2005,
  title = {The {{Smallest Grammar Problem}}},
  author = {Charikar, M. and Lehman, E. and Liu, D. and Panigrahy, R. and Prabhakaran, M. and Sahai, A. and Shelat, A.},
  date = {2005-07},
year={2005},
  journal = {IEEE Transactions on Information Theory},
  shortjournal = {IEEE Trans. Inform. Theory},
  volume = {51},
  number = {7},
  pages = {2554--2576},
  issn = {0018-9448},
  doi = {10.1109/TIT.2005.850116},
  url = {http://ieeexplore.ieee.org/document/1459058/},
  urldate = {2025-09-19},
  langid = {english}
}

@article{xu2020numeral,
  title={Numeral systems across languages support efficient communication: From approximate numerosity to recursion},
  author={Xu, Yang and Liu, Emmy and Regier, Terry},
  journal={Open Mind},
  volume={4},
  pages={57--70},
  year={2020},
url={https://doi.org/10.1162/opmi_a_00034}
}

@phdthesis{brighton2003simplicity,
  title        = {Simplicity as a Driving Force in Linguistic Evolution},
  author       = {Brighton, Henry},
  year         = {2003},
  school       = {The University of Edinburgh},
  type         = {PhD thesis},
  url          = {http://hdl.handle.net/1842/23810}
}

@article{dehaene1992cross,
  title={Cross-linguistic regularities in the frequency of number words},
  author={Dehaene, Stanislas and Mehler, Jacques},
  journal={Cognition},
  volume={43},
  number={1},
  pages={1--29},
  year={1992},
  publisher={Elsevier},
url={https://doi.org/10.1016/0010-0277(92)90030-L}
}

@book{Hurford1975TheLT,
  title={The linguistic theory of numerals},
  author={Hurford, James},
  volume={16},
  year={1975},
  publisher={Cambridge University Press}
}

@article{HURFORD2007773,
title = {A performed practice explains a linguistic universal: Counting gives the Packing Strategy},
journal = {Lingua},
volume = {117},
number = {5},
pages = {773-783},
year = {2007},
note = {Linguistic perspectives on numerical expressions},
issn = {0024-3841},
url = {https://doi.org/10.1016/j.lingua.2006.03.002},
author = {James Hurford}
}

@incollection{wals-s1,
  author    = {Bernard Comrie and Matthew S. Dryer and David Gil and Martin Haspelmath},
  booktitle = {The World Atlas of Language Structures Online},
  editor    = {Matthew S. Dryer and Martin Haspelmath},
  publisher = {Zenodo},
  title     = {Introduction (v2020.4)},
  type      = {Data set},
  url       = {https://doi.org/10.5281/zenodo.13950591},
  year      = {2013},
  doi       = {10.5281/zenodo.13950591}
}

@inproceedings{carlsson2021learning,
  title={Learning approximate and exact numeral systems via reinforcement learning},
  author={Carlsson, Emil and Dubhashi, Devdatt and Johansson, Fredrik D},
  booktitle={Proceedings of the Annual Meeting of the Cognitive Science Society},
 volume ={43},
  year={2021}
}

@book{mihovFiniteStateTechniquesAutomata2019,
  title = {Finite-{{State Techniques}}: {{Automata}}, {{Transducers}} and {{Bimachines}}},
  shorttitle = {Finite-{{State Techniques}}},
  author = {Mihov, Stoyan and Schulz, Klaus U.},
  date = {2019-08-01},
year={2019},
  edition = {1},
  publisher = {Cambridge University Press},
  doi = {10.1017/9781108756945},
  urldate = {2025-09-24},
  isbn = {978-1-108-75694-5 978-1-108-48541-8}
}

@article{regier2015word,
  title={Word meanings across languages support efficient communication},
  author={Regier, Terry and Kemp, Charles and Kay, Paul},
  journal={The handbook of language emergence},
  pages={237--263},
  year={2015},
  publisher={Wiley Online Library},
  url={https://doi.org/10.1002/9781118346136.ch11}
}

@article{zaslavsky2019col,
author = {Zaslavsky, Noga and Kemp, Charles and Tishby, Naftali and Regier, Terry},
title = {Color Naming Reflects Both Perceptual Structure and Communicative Need},
journal = {Topics in Cognitive Science},
volume = {11},
number = {1},
pages = {207-219},
doi = {https://doi.org/10.1111/tops.12395},
url = {https://doi.org/10.1111/tops.12395},
year = {2019}
}

@article{uegaki_informativenesscomplexity_2022,
    title = {The informativeness/complexity trade-off in the domain of boolean connectives},
    issn = {0024-3892, 1530-9150},
    url = {https://direct.mit.edu/ling/article/doi/10.1162/ling_a_00461/109419/The-Informativeness-Complexity-Trade-Off-in-the},
    doi = {10.1162/ling_a_00461},
    language = {en},
    urldate = {2022-12-21},
    journal = {Linguistic Inquiry},
    author = {Uegaki, Wataru},
    month = oct,
    year = {2022},
    pages = {1--23},
}

@article{zaslavsky2018efficient,
  title={Efficient compression in color naming and its evolution},
  author={Zaslavsky, Noga and Kemp, Charles and Regier, Terry and Tishby, Naftali},
  journal={Proceedings of the National Academy of Sciences},
  volume={115},
  number={31},
  pages={7937--7942},
  year={2018},
  publisher={National Academy of Sciences}
}

@article{culbertson2016simplicity,
  title={Simplicity and specificity in language: Domain-general biases have domain-specific effects},
  author={Culbertson, Jennifer and Kirby, Simon},
  journal={Frontiers in psychology},
  volume={6},
  pages={1964},
  year={2016},
  publisher={Frontiers Media SA}
}

@article{bentz2023complexity,
url = {https://doi.org/10.1515/lingvan-2021-0054},
title = {Complexity trade-offs and equi-complexity in natural languages: a meta-analysis},
author = {Christian Bentz and Ximena Gutierrez-Vasques and Olga Sozinova and Tanja Samardžić},
pages = {9--25},
volume = {9},
number = {s1},
journal = {Linguistics Vanguard},
doi = {10.1515/lingvan-2021-0054},
year = {2023},
lastchecked = {2026-01-23}
}

@article{culbertson2012learning,
  title={Learning biases predict a word order universal},
  author={Culbertson, Jennifer and Smolensky, Paul and Legendre, G{\'e}raldine},
  journal={Cognition},
  volume={122},
  number={3},
  pages={306--329},
  year={2012},
  publisher={Elsevier}
}

@article{kirby2015compression,
  title={Compression and communication in the cultural evolution of linguistic structure},
  author={Kirby, Simon and Tamariz, Monica and Cornish, Hannah and Smith, Kenny},
  journal={Cognition},
  volume={141},
  pages={87--102},
  year={2015},
  publisher={Elsevier},
  url={https://doi.org/10.1016/j.cognition.2015.03.016}
}

@article{carrSimplicityInformativenessSemantic2020,
  title = {Simplicity and Informativeness in Semantic Category Systems},
  author = {Carr, Jon W. and Smith, Kenny and Culbertson, Jennifer and Kirby, Simon},
  year = 2020,
  month = sep,
  journal = {Cognition},
  volume = {202},
  pages = {104289},
  issn = {00100277},
  doi = {10.1016/j.cognition.2020.104289},
  url = {https://doi.org/10.1016/j.cognition.2020.104289},
  urldate = {2023-06-12},
  langid = {english},
}

@article{van2023quantifiers,
  title={Quantifiers satisfying semantic universals have shorter minimal description length},
  author={van de Pol, Iris and Lodder, Paul and van Maanen, Leendert and Steinert-Threlkeld, Shane and Szymanik, Jakub},
  journal={Cognition},
  volume={232},
  pages={105150},
  year={2023},
  publisher={Elsevier}
}

@article{prasertsom2026domain,
  title={Domain-general categorisation explains constrained cross-linguistic variation in noun classification},
  author={Prasertsom, Ponrawee and Smith, Kenny and Culbertson, Jennifer},
  journal={Cognition},
  volume={271},
  pages={106411},
  year={2026},
  url={https://doi.org/10.1016/j.cognition.2025.106411},
  publisher={Elsevier}
}

@article{STEINERTTHRELKELD2020104076,
title = {Ease of learning explains semantic universals},
journal = {Cognition},
volume = {195},
pages = {104076},
year = {2020},
issn = {0010-0277},
doi = {https://doi.org/10.1016/j.cognition.2019.104076},
url = {https://www.sciencedirect.com/science/article/pii/S0010027719302495},
author = {Shane Steinert-Threlkeld and Jakub Szymanik}
}
\bibliographystyle{acl_natbib}

\appendix
\SetKwInput{KwData}{Requires}
\section{Genetic Algorithm of \citet{denic2024recursive}}
\label{sec:algorithms}

In this Appendix we describe the Genetic Algorithm used by \citet{denic2024recursive} to generate the Pareto frontier of optimal recursive numeral systems (Algorithm \ref{alg:ga}). Note that D\&S to evaluate a $(D,M)$ pair utilise Hurford's grammar to generate numerals for each number from $1-99$, keeping the shortest numeral for each of them. \citet{yang2025re} utilise a similar algorithm but includes also suppletives $S$ beside digits and multipliers, use the constrains over Hurford's grammar discussed in Section \ref{sec:literature}. While D\&S perform mutations only on the sets $D$ and $M$, Y\&R includes the suppletives $S$ and exception rules.

\begin{algorithm}[hbt!]
\caption{Genetic algorithm for Pareto frontier.}\label{alg:ga}
\KwResult{Pareto-optimal population of $(D,M)$ pairs.}
Sample initial population of $(D,M)$ pairs\;

Evaluate population (via lexicon size and average morphosyntactic complexity)\; 
\For{$i \in [1, \dots, \text{max generation}]$}
{
    Select only Pareto-dominant $(D,M)$ pairs\;
    Perform 1 to 3 random mutations\;
    Evaluate dominant systems and offspring\;
}
\end{algorithm}

\section{Estimating natural languages' local neighbourhood frontiers}\label{sec:appendix_control}

\begin{algorithm}[hbt!]
\caption{Local neighbourhood frontier estimation.}\label{alg:control}

\KwData{digits $D$, multipliers $M$, combinators $C$, numeral lengths $L_{Num}$; max sample size $\beta$, expansion size $\gamma$, depth $d$.}

\KwResult{The set of local neighbourhood's optimal systems $\mathcal{O}$.}

\For{$i \in [1, \dots, 99]$}
{   
    Generate numerals of number $i$ up to depth $d$;
    
    Keep only the numeral(s) of $i$ whose length matches $L_{Num}[i]$;
}

$N_0=$ \{numerals with no alternative form\};

$\mathcal{O}=\{N_0\}$\;

\While{$|\mathcal{O}[i]|<99 \; \forall i \in \{1,\; ..., \;|\mathcal{O}|\}$}
{   
    Enumerate in $\mathcal{S}$ the combinations of the next $\gamma$ numbers' alternative numerals\;
    \For{optimal sub-system $O$ in $\mathcal{O}$}
    {

    \For{set of alternatives $S$ in $\mathcal{S}$}
    {
        Calculate $L(G)$ and $L(N \mid G)$ of $O \cup S$.
    }
    }

    $\mathcal{O}=\{O\cup S : O\cup S \textnormal{ is Pareto-dominant}\}$;

    \If{$|\mathcal{O}|> \beta$}
    {
        Sample $\beta$ systems from $\mathcal{O}$.
    
    }
}
\end{algorithm}

Algorithm \ref{alg:control} (Section \ref{sec:control}) approximates the local neighbourhood of natural recursive numeral systems. 
It works as follows.
For a given local neighbourhood $(D,M,C,L_{Num})$, the algorithm works as follows.
First, we fix all the numerals that have no alternative form other than the real one (e.g., Mandarin $10+2$ only has one alternative of length 3, namely $10+2$).
From this initial partial system, we enumerate all the possible partial systems that have all the initial numerals, plus the combination of all numerals for the next $\gamma{}$ largest numbers.

We identify the Pareto-optimal partial systems in terms of our measures and discarded the rest.
If the number of Pareto-optimal partial systems exceeded $\beta{}$, we sample only $\beta{}$ systems.
For each of these $\beta{}$ systems, we enumerate again all the possible partial systems that included the combinations of numerals for the next $\gamma{}$ numbers.
We repeat this process until we obtained the full systems that can express the numbers from 1 to 99.

Note that when we are searching for the least efficient systems in the local neighbourhood, we keep in $\mathcal{O}$ only the sub-systems that are Pareto-optimal for the \textit{negative} of our measures.

We run the algorithm with 37 out of 38 recursive numeral system types inferred from the D\&S sample of natural systems.
We exclude the Kayah Li-like type because they require a search depth deeper than our max depth for this task ($d = 6$) to generate a space that includes Kayah Li itself.
Out of the 37 types, we set $\beta{} = 30$ and $\gamma{} = 3$ for 30 of them.
We set $\beta{} = 10, \gamma{}= 2$ for the other 7 language types (Kilivila-, Abkhaz-, Gola-, Otomi-, Supyire-, Khana-, and Mangab-Mbula-like), for which sampling is too computationally expensive. 

\section{Correlation between measures in random baseline and natural systems}
\label{sec:correlation}

In this Appendix, we report Pearson's correlations between regularity, processing complexity and average morphosyntactic complexity. 
As we are not able to sample the whole space of possible recursive numeral systems,
our analysis is limited to the 10,000 baseline recursive numeral systems we randomly generate following the procedure described in Section \ref{sec:langcomparison} (orange dots in Figure \ref{fig:randomsamplecomp}).

Table \ref{tab:corr-matrix-random} suggests that all three measure positively correlate.
While this seems to suggest that these measures (including regularity and processing complexity) are not independent, we note that it is likely because our space is not completely random, as it is limited by our computational resource restrictions as well as how the samples are obtained.
We only consider systems with 3-12 digits, 1-3 multipliers, and only use attested digits/multipliers.
This method, together with the fact that numerals have to evaluate correctly to their corresponding numbers in the range 1--99, likely imposes  additional constraints on the shape of the space.  The correlations reported here thus likely do not reflect how these measures would correlate if we were able to sample from the whole space of possible recursive numeral systems.

\begin{table}[!ht]
\centering
\resizebox{\columnwidth}{!}{%
\begin{tabular}{lccc}
\hline
 & avg. morpho. & (ir)regularity & proc. complexity \\
\hline
avg. morpho.& - & 0.659 & 0.873 \\
(ir)regularity                  & 0.659 & - & 0.625 \\
proc. complexity           & 0.873 & 0.625 & - \\
\hline
\end{tabular}%
}
\caption{Correlation matrix of random baseline systems.}
\label{tab:corr-matrix-random}
\end{table}

Importantly, the correlations seem to be introduced because of the sampled systems that are relatively efficient (the bulk of random systems that have around 4-6 bits of processing complexity in Figure \ref{fig:randomsamplecomp}).
In fact, if we consider the space of randomly sampled recursive numeral systems that have processing complexity above 6 bits (which correspond to 6,468 systems, roughly 65\% of all systems) then the correlations are significantly weaker (Table \ref{tab:corr-matrix-random-base10}). 

\begin{table}[!ht]
\centering
\resizebox{\columnwidth}{!}{%
\begin{tabular}{lccc}
\hline
& avg. morpho. & (ir)regularity & proc. complexity \\
\hline
avg. morpho. & - & 0.186 & 0.246 \\
(ir)regularity                  & 0.186 & - & 0.129 \\
proc. complexity           & 0.246 & 0.129 & - \\
\hline
\end{tabular}%
}
\caption{Correlation matrix of random baseline languages with processing complexity greater than 6 bits.}
\label{tab:corr-matrix-random-base10}
\end{table}

\end{document}